# Estimating Well-Performing Bayesian Networks using Bernoulli Mixtures

**Geoff Jarrad**

The Business Intelligence Group
CSIRO Mathematical and Information Sciences
PMB 2, Glen Osmond, South Australia 5064.
e-mail: Geoff.Jarrad@cmis.csiro.au

## Abstract

A novel method for estimating Bayesian network (BN) parameters from data is presented which provides improved performance on test data. Previous research has shown the value of representing conditional probability distributions (CPDs) via neural networks (Neal 1992), noisy-OR gates (Neal 1992, Diez 1993) and decision trees (Friedman and Goldszmidt 1996). The Bernoulli mixture network (BMN) explicitly represents the CPDs of discrete BN nodes as mixtures of local distributions, each having a different set of parents. This increases the space of possible structures which can be considered, enabling the CPDs to have finer-grained dependencies. The resulting estimation procedure induces a model that is better able to emulate the underlying interactions occurring in the data than conventional conditional Bernoulli network models. The results for artificially generated data indicate that overfitting is best reduced by restricting the complexity of candidate mixture substructures local to each node. Furthermore, mixtures of very simple substructures can perform almost as well as more complex ones. The BMN is also applied to data collected from an online adventure game with an application to keyhole plan recognition. The results show that the BMN-based model brings a dramatic improvement in performance over a conventional conditional Bernoulli BN model.

## 1 INTRODUCTION

The problem of estimating the parameters and structure of a Bayesian network (BN) from prior information and observed data has received a great deal of attention in recent years (Buntine 1996 gives a review of the literature). A common approach for structure estimation is some form of Bayesian model selection (BMS), which typically selects the single BN model (or an equivalence class of such models) with the maximum likelihood (ML). Alternatively, prior weights can be chosen to penalise complex structures, and the BN model with the the maximum *a posteriori* (MAP) likelihood can be found instead (Heckerman 1995). The BMS approach is reasonable when there is a large amount of data and the BN has a small number of nodes (Cooper and Herskovits 1992), but this condition often does not hold true (Friedman and Koller 2000), in which case the model chosen by BMS might not be representative of the underlying processes. That is, the selected model will often overfit the data and not generalise to new data.

An alternative to BMS is to average over many BNs with different structures, using a mixture of Bayesian networks (an MBN — Thiesson *et al.* 1997), Typically, the BN parameters are integrated out to form the marginal likelihood of the data for each BN structure. Bayesian model averaging (BMA) takes this a step further by summing over the individual BN structures, which gives better average predictions (Hoeting *et al.* 1999). The difficulty with such approaches is that the marginal likelihood is usually difficult to compute. In particular, when data are missing, or when mixtures are used, the marginal likelihood does not have a closed form and a large-sample approximation (Kass *et al.* 1988) is often used. Such approximations, however, typically require very large amounts of data to be accurate.

The method proposed in this paper is not concerned with structure estimation *per se*, but rather with the use of structure to obtain good estimates for the conditional probability distributions (CPDs) of a given Bayesian network. This single BN could be obtained from BMS, or could be specified in advance by a domain expert. The **Bernoulli[1] mixture network** (BMN) model (which is a discrete version of the more general Bayesian mixture-network model) is based on a novel idea of McMichael (1998), in which the CPD of each node in a BN is treated as a mixture of local distributions, each having a different subset of parents. Previous approaches to using local representations of CPDs have included neural networks (Neal 1992), noisy-OR gates (Neal 1992, Diez 1993) and decision trees (Friedman and Goldszmidt 1996). These approaches, including the BMN, require the CPDs to be estimated from the data, and hence they differ from methods which approximate known CPDs by other distributions (e.g. Tresp *et al.* 1999).

There is also a special relationship between the BMN and the MBN. Recall that the MBN averages over the joint distributions of global BN structures for various node orderings, whereas the BMN averages over CPDs of local structures for a single BN. This means that the MBN can in general span a much larger space of network structures than the BMN. However, an MBN restricted to one particular ordering reduces to an equivalent BMN (see Section 2.3). Empirical evidence has verified that the performances of the restricted MBN and the BMN on complete data are the same. As a consequence, the BMN, which is stored as a single BN, is more efficient than the corresponding MBN, which is stored as a collection of BNs with some duplication of parameter estimates.

# 2 THE MIXTURE MODEL

## 2.1 BACKGROUND

A Bayesian network (BN) is a directed acyclic graph (DAG) $\mathcal{G} = \{\mathcal{V}, \mathcal{S}\}$, with vertex set $\mathcal{V}$ and acyclic structure $\mathcal{S}$, coupled with a collection $\mathbf{\Theta}$ of parameters. It is assumed that $\mathcal{G}$ has a particular node ordering $\mathcal{I} = \{1, 2, \ldots, V\}$, with each node (or vertex or variable) labelled as $\nu_i \in \mathcal{V}$ for index $i \in \mathcal{I}$. Under the Markov property, $\mathcal{G}$ is decomposable in the sense that each node $\nu_i$ depends only upon its set of parents $\mathcal{P}_i = \{\nu_j \mid \nu_j \to \nu_i\}$, where $\nu_j \to \nu_i$ denotes a directed arc from node $\nu_j$ to $\nu_i$. This in turn defines the local structure $\widetilde{\mathcal{S}}_i = \{j \in \mathcal{I} \mid \nu_j \in \mathcal{P}_i\}$ for each node $\nu_i$, so that the overall structure is $\mathcal{S} = (\widetilde{\mathcal{S}}_i)_{i=1}^V$. A similar decomposibility $\mathbf{\Theta} = (\Theta_i)_{i=1}^V$ is assumed for the BN parameters, where each node $\nu_i$ has the local parameter $\Theta_i$.

Given these definitions, the likelihood of the joint state of the BN is

$$P(\boldsymbol{\nu}|\mathbf{\Theta}, \mathcal{S}) \quad = \quad \prod_{i=1}^{V} P(\nu_i|\pi_i, \Theta_i)\,, \tag{1}$$

where, in a slight abuse of notation, $\pi_i$ here represents both the ordered list of nodes in $\mathcal{P}_i$, and the arbitrary values of those nodes. This BN model is known as the *conventional model*, in order to distinguish it from the varieties of mixture models presented in the next two sections. For a discrete BN, in which the CPD of each node is multinomial, equation (1) further reduces to

$$P(\boldsymbol{\nu}|\mathbf{\Theta}, \mathcal{S}) \quad = \quad \prod_{i=1}^{V} \theta_{i,\nu_i,\pi_i}\,, \tag{2}$$

where $\Theta_i = [[\theta_{ijk}]_{j=1}^{Q_i}]_{k=1}^{R_i}$. For convenience, the conditional probability table (CPT) of the $i$th node will simply be represented from now on as $\Theta_i = [\theta_{ijk}]$.

## 2.2 A LOCAL MIXTURE MODEL

The *Bayesian mixture-network* model is formed from the conventional model, of the previous section, by decomposing the local structure $\widetilde{\mathcal{S}}_i$ into a list of $M_i$ candidate substructures, namely $\mathcal{S}_i = (\mathcal{S}_{i,m_i})_{m_i=1}^{M_i}$, so that $\mathcal{S} = (\mathcal{S}_i)_{i=1}^V$. Observe that in this slightly altered notation, the conventional model is now equivalent to a mixture model with a singleton list $\mathcal{S}_i = (\widetilde{\mathcal{S}}_i)$ of substructures for each node $\nu_i$, i.e. $M_i \equiv 1$. In general, for each mixture component $m_i = 1, 2, \ldots, M_i$, the substructure $\mathcal{S}_{i,m_i} \subseteq \widetilde{\mathcal{S}}_i$ has the corresponding BN parameter $\Theta_{i,m_i}$, so that $\mathbf{\Theta}_i = (\Theta_{i,m_i})_{m_i=1}^{M_i}$ and $\mathbf{\Theta} = (\mathbf{\Theta}_i)_{i=1}^V$. The mixture weights are parameterised by $\mathbf{\Psi} = (\mathbf{\Psi}_i)_{i=1}^V$, where $\mathbf{\Psi}_i = (\psi_{i,m_i})_{m_i=1}^{M_i}$. The likelihood of the joint state of the BN for this mixture model is

$$P(\boldsymbol{\nu}|\mathbf{\Psi}, \mathbf{\Theta}, \mathcal{S}) \quad = \quad \prod_{i=1}^{V} P(\nu_i|\pi_i, \mathbf{\Psi}_i, \mathbf{\Theta}_i)\,, \tag{3}$$

where

$$P(\nu_i|\pi_i, \mathbf{\Psi}_i, \mathbf{\Theta}_i) \quad = \quad \sum_{m_i=1}^{M_i} P(m_i|\pi_i, \mathbf{\Psi}_i)\; P(\nu_i|\pi_i, \mathbf{\Theta}_i, m_i)\,. \tag{4}$$

The mixture weights are further assumed to depend only upon the candidate substructures, and not on the actual states of the parents. Hence, the Bayesian mixture-network model is given by

$$P(\boldsymbol{\nu}|\mathbf{\Psi}, \mathbf{\Theta}, \mathcal{S}) \quad = \quad \prod_{i=1}^{V} \sum_{m_i=1}^{M_i} P(m_i|\mathbf{\Psi}_i)\; P(\nu_i|\pi_i, \Theta_{i,m_i})\,. \tag{5}$$

[1] The term "Bernoulli" is used here in the n-ary or multinomial sense.

The discrete form of this model is the *Bernoulli mixture network* (BMN) model:

$$P(\nu|\Psi,\Theta,\mathcal{S}) = \prod_{i=1}^{V}\sum_{m_i=1}^{M_i}\psi_{i,m_i}\,\theta_{i,\nu_i,\pi_i,m_i}\,, \quad (6)$$

which is parameterised by:

$$P(m_i|\Psi_i) = \psi_{i,m_i}\,, \quad (7)$$

$$P(\nu_i|\pi_i,\Theta_{i,m_i}) = \theta_{i,\nu_i,\pi_i,m_i}\,, \quad (8)$$

where the CPT for the $m_i$th substructure of the $i$th node is $\Theta_{i,m_i} = [\theta_{ijk,m_i}]$.

The total number of possible substructures for node $\nu_i$ is given by $M_i = 2^{P_i}$, where there are $P_i = |\mathcal{P}_i|$ parents in the conventional BN model, and any given substructure can have zero, one, two or more of the candidate parents selected. Observe that for the given node ordering $\mathcal{I}$, the $i$th node can have a maximum of $i-1$ parents (i.e. $P_i \leq i-1$). The special substructure in which all $i-1$ candidate parents are selected is known as the `full` substructure. Correspondingly, the BN structure with full substructure for every node is called the `full` structure.

It is desirable, however, to limit the number of selected parents in each substructure in order to reduce overfitting (see Section 4.1). Thus, if at most $p_i \leq P_i$ parents are allowed to be selected from the candidate set, then

$$M_i = \sum_{\tilde{p}=0}^{p_i}\binom{P_i}{\tilde{p}} \leq 2^{P_i}\,, \quad (9)$$

where equality holds only if $p_i = P_i$. Typically $p_i$ might be roughly $P_i/2$, which is the expected number of parents if any given directed arc has equal probability of being present or absent.

### 2.3 A GLOBAL MIXTURE MODEL

In contrast to the Bayesian mixture model of the previous section, the *mixture of Bayesian networks* (MBN) model averages over a selection of global BN structures, possibly for a variety of node orderings. In particular, the overall collection of BN parameters has the decomposition $\Theta = (\Theta_m)_{m=1}^{M}$ for $\Theta_m = (\Theta_{m,i})_{i=1}^{V}$, where $\Theta_{m,i}$ parameterises the CPD of the $i$th node in the $m$th structure. Here $M$ is the total number of candidate structures. Likewise, the mixture parameters are decomposed as $\Psi = (\Psi_m)_{m=1}^{M}$. Hence, the likelihood of the joint state of all nodes is

$$P(\nu|\Psi,\Theta,\mathcal{S}) = \sum_{m=1}^{M} P(m|\Psi)\prod_{i=1}^{V}P(\nu_i|\pi_i,\Theta_{m,i})\,. \quad (10)$$

The discrete form of this MBN is then just

$$P(\nu|\Psi,\Theta,\mathcal{S}) = \sum_{m=1}^{M}\Psi_m\prod_{i=1}^{V}\theta_{m,i,\nu_i,\pi_i}\,, \quad (11)$$

where

$$P(m|\Psi) = \Psi_m\,, \quad (12)$$

$$P(\nu_i|\pi_i,\Theta_{m,i}) = \theta_{m,i,\nu_i,\pi_i}\,, \quad (13)$$

and $\Theta_{m,i} = [\theta_{m,ijk}]$ is the appropriate CPT.

Observe that this MBN has some resemblence to the BMN of the previous section. In fact, the MBN is a generalisation of the BMN. To see this, the explicit node ordering $\mathcal{I}$ is first imposed, giving $M = 2^{V(V-1)/2}$ candidate structures. Next, the mixture weight $\Psi_m$ of the $m$th structure is decomposed on a node-by-node basis as $\Psi_m = \prod_{i=1}^{V}\psi_{m,i}$. Equation (11) then becomes

$$P(\nu|\Psi,\Theta,\mathcal{S}) = \sum_{m=1}^{2^{V(V-1)/2}}\prod_{i=1}^{V}\psi_{m,i}\,\theta_{m,i,\nu_i,\pi_i}$$

$$= \prod_{i=1}^{V}\sum_{m_i=1}^{2^{i-1}}\psi_{\bar{m}(i,m_i),i}\,\theta_{\bar{m}(i,m_i),i,\nu_i,\pi_i}\,, \quad (14)$$

where each pair $(i, m_i)$ is mapped by $\bar{m}$ to a unique $m$. Thus, by equating $\psi_{m,i}$ with $\psi_{i,m_i}$ and $\theta_{m,i,\nu_i,\pi_i}$ with $\theta_{i,\nu_i,\pi_i,m_i}$, it can be seen that for complete data the MBN with restricted ordering is equivalent to the BMN given by equation (6), where $M_i = 2^{i-1}$. However, the BMN representation is more compact in the sense that it can span the $M = 2^{V(V-1)/2}$ global structures using only $\sum_{i=1}^{V} M_i = 2^V - 1$ local mixture components, which can be collapsed to $V$ parameter estimates.

## 3 PARAMETER ESTIMATION

Parameters for both the conventional BN model and the BMN are estimated from a collection $\boldsymbol{X} = (\boldsymbol{x}_d)_{d=1}^{N}$ of $N$ observed cases, known as the training data. Missing data are handled using a collection $\boldsymbol{Z}$ of uniformly distributed indicator variables, in conjunction with the expectation-maximisation (EM) algorithm (Dempster *et al.* 1977), which has known monotonic convergence (Boyles 1983, and Wu 1983). The conditional probabilities are then obtained using maximum *a posteriori* (MAP) estimation (McMichael 1998), as described briefly below.

Recall that the BMN model is given by equation (6). Thus, allowing for missing data, the likelihood of the $d$th case is

$$P(\boldsymbol{x}_d|\Psi,\Theta,\mathcal{S},Z) = \prod_{i=1}^{V}\prod_{j=1}^{Q_i}\prod_{k=1}^{R_i}\prod_{m_i=1}^{M_i} \psi_{i,m_i}^{z_{di,m_i}} \theta_{ijk,m_i}^{z_{dijk,m_i}}, \quad (15)$$

where $z_{di,m_i} = 1$ if submodel $m_i$ is the correct choice at the $i$th node for the $d$th observed case, and $z_{dijk,m_i} = 1$ if, further, the family $(\nu_i, \pi_i)$ is in state $(j,k)$; otherwise $z_{di,m_i} = 0$ and $z_{dijk,m_i} = 0$. The expected log-likelihood is then

$$\begin{aligned} Q(\Theta,\Psi,\Theta',\Psi') &= \mathcal{E}_{\mathbf{Z}}\left[\log P(\boldsymbol{X}|\Psi,\Theta,\mathcal{S},\boldsymbol{Z}) \mid \boldsymbol{X},\Theta',\Psi'\right] \\ &= \sum_{i=1}^{V}\sum_{j=1}^{Q_i}\sum_{k=1}^{R_i}\sum_{m_i=1}^{M_i} [N_{ijk,m_i}\log\theta_{ijk,m_i} \\ &\qquad + A_{i,m_i}\log\psi_{i,m_i}], \quad (16) \end{aligned}$$

where

$$\begin{aligned} N_{ijk,m_i} &= \sum_{d=1}^{N} P(\nu_i = j, \pi_i = k|\boldsymbol{x}_d, \Theta', \Psi', m_i) \\ &\qquad \times P(m_i|\boldsymbol{x}_d, \Theta', \Psi'), \quad (17) \\ A_{i,m_i} &= \sum_{d=1}^{N} P(m_i|\boldsymbol{x}_d, \Theta', \Psi'). \quad (18) \end{aligned}$$

The MAP objective function $F$ now combines the expected log-likelihood with joint priors for the parameters, along with a condition $h(\Psi)$ to ensure the mixture weights sum to unity. The prior density for $\Theta$ is

$$p(\Theta|\mathcal{S}) = a\prod_{i=1}^{V}\prod_{j=1}^{Q_i}\prod_{k=1}^{R_i}\prod_{m_i=1}^{M_i} \theta_{ijk,m_i}^{N^0_{ijk,m_i}}, \quad (19)$$

for normalising constant $a$, and the prior density for the mixture weights $\Psi$ is

$$p(\Psi|\mathcal{S}) = b\prod_{i=1}^{V}\prod_{m_i=1}^{M_i} \psi_{i,m_i}^{\alpha_{i,m_i}}, \quad (20)$$

for normalising constant $b$. Thus, the objective function is

$$\begin{aligned} F(\Theta,\Psi,\Theta',\Psi') &= Q(\Theta,\Psi,\Theta',\Psi') + g(\Theta) + h(\Psi) \\ &\quad + \log p(\Theta|\mathcal{S}) + \log p(\Psi|\mathcal{S}) \\ &= \sum_{i=1}^{V}\sum_{j=1}^{Q_i}\sum_{k=1}^{R_i}\sum_{m_i=1}^{M_i} \\ &\quad [(N^0_{ijk,m_i} + N_{ijk,m_i})\log\theta_{ijk,m_i} \\ &\quad + (\alpha_{i,m_i} + A_{i,m_i})\log\psi_{i,m_i}] \\ &\quad + \sum_{i=1}^{V}\sum_{k=1}^{R_i}\sum_{m_i=1}^{M_i} \lambda_{ik,m_i}\left[\sum_{j=1}^{Q_i}\theta_{ijk,m_i} - 1\right] \\ &\quad + \sum_{i=1}^{V}\mu_i\left[\sum_{m_i=1}^{M_i}\psi_{i,m_i} - 1\right], \quad (21) \end{aligned}$$

where the normalising constants $a$ and $b$ have been neglected. The MAP estimates are then found from the local maximum conditions $\frac{\partial F}{\partial\theta_{ijk,m_i}} = 0$ and $\frac{\partial F}{\partial\psi_{i,m_i}} = 0$ to be

$$\begin{aligned} \theta_{ijk,m_i} &= \frac{N^0_{ijk,m_i} + N_{ijk,m_i}}{N^0_{i\cdot k,m_i} + N_{i\cdot k,m_i}}, \quad (22) \\ \psi_{i,m_i} &= \frac{\alpha_{i,m_i} + A_{i,m_i}}{\alpha_{i,\cdot} + N}, \quad (23) \end{aligned}$$

where $N_{i\cdot k,m_i} = \sum_{j=1}^{Q_i} N_{ijk,m_i}$ (and likewise for $N^0_{i\cdot k,m_i}$), and $\alpha_{i,\cdot} = \sum_{m_i=1}^{M_i}\alpha_{i,m_i}$.

There are two problems with this formulation. The first is that an expert can reasonably be expected to specify the prior counts $N^0_{ijk}$, but can not be expected to specify the mixture counts $N^0_{ijk,m_i}$, since there will in general be too many submodel interactions with which to cope. The second problem is that equation (17) for $N_{ijk,m_i}$ is both ambiguous and intractible, for much the same reason. For instance, equation (17) specifies the use of the $m_i$th submodel at the $i$th node, but does not specify which conditional probabilities are to be used at other nodes.

These problems can both be solved by following the design philosophy that the BMN reduces to a conventional BN after parameter estimation. Thus, each submodel count $N_{ijk,m_i}$ is computed from the table $[N_{ijk}]$ of family data counts by summing over the states of all parents not included in the $m_i$th submodel. Likewise, each $N^0_{ijk,m_i}$ can be computed from table $[N^0_{ijk}]$ of prior family counts supplied by the expert. The family data counts themselves are computed from the conventional BN containing all candidate parents, via

$$N_{ijk} = \sum_{d=1}^{N} P(\nu_i = j, \pi_i = k|\boldsymbol{x}_d, \bar{\Theta}), \quad (24)$$

where $\bar{\Theta} = (\bar{\Theta}_i)_{i=1}^{V}$, $\bar{\Theta}_i = [\bar{\theta}_{ijk}]$ and

$$\begin{aligned} \bar{\theta}_{ijk} &= P(\nu_i = j|\pi_i = k, \bar{\Theta}_i) \\ &= \sum_{m_i=1}^{M_i} P(m_i|\Psi_i)\, P(\nu_i = j|\pi_i = k, \Theta_{i,m_i}) \\ &= \sum_{m_i=1}^{M_i} \psi_{i,m_i}\theta_{ijk,m_i}. \quad (25) \end{aligned}$$

# 4 EMPIRICAL RESULTS

## 4.1 ARTIFICIAL EXPERIMENT

The following experiment was designed to test the feasibility of recovering the BN model with known structure solely from the data it generates. The performance of the BMN with various restrictions on the submodels is compared to the usual method of model selection. This reveals some of the more apparent properties of the BMN, and is not intended to systematically address all issues. Further experiments are required to show more subtle properties, such as the dependence on the number of nodes and number of local mixtures.

The artificially-created model (the so-called `true` model) chosen for this experiment has the network shown in Figure 1. The $V = 4$ nodes `One`, `Two`, `Three` and `Four` have 3, 2, 2 and 3 discrete states, respectively. The conditional probability tables were randomly initialised, and the resulting `true` model was used to randomly generate $N_{\mathtt{train}} = 100$ complete cases of training data and $N_{\mathtt{test}} = 2000$ complete cases of testing data.

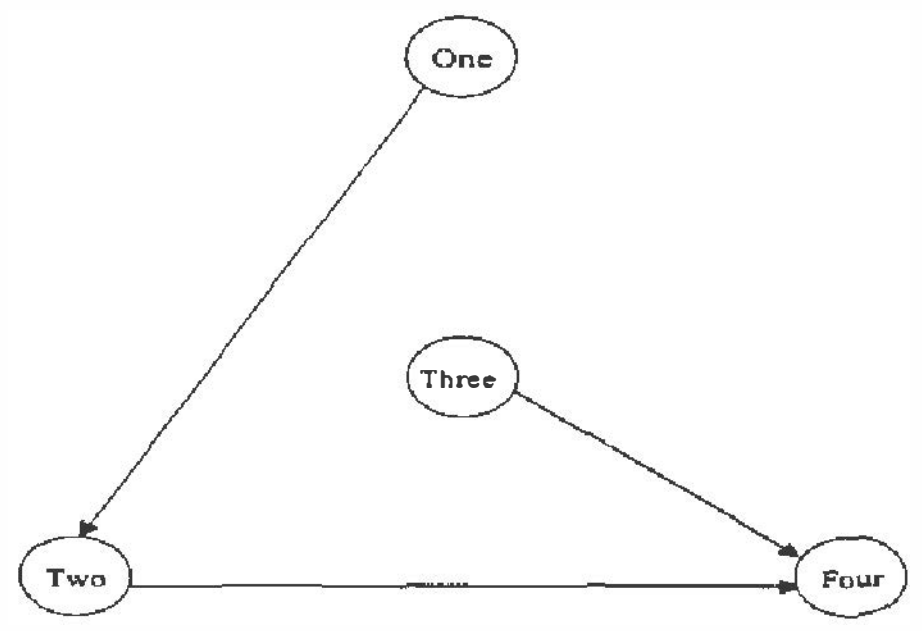

Figure 1: The `true` model structure

The relative performances of the $V! \times 2^{V(V-1)/2} = 1536$ different possible BN models on the training and testing data are shown in Figure 2, using the log-likelihood divided by the number of data as a normalised score. Of special interest are the `empty` model (1) with completely independent nodes (i.e. no edges), the `true` model (2), and the `full` model (3) with completely dependent nodes. The `full` BN model (or the equivalence class of $V! = 24$ such models for all nodes orderings) is special in that it embodies the exact rule of conditional probability, and therefore fits the complete training data exactly. Thus, Bayesian model selection (Ramoni and Sebastiani 1997) with uniform prior model weights results in the maximum-likelihood choice (4) of the `full` model, shown in Figure 3.

Observe from Figure 2 that there is a band of models, including the `full` model (1), which fit the training data well and the testing data poorly, indicating overfitting. There is also another band of better performing models with roughly linear correspondence between the two scores, with the `empty` BN (1) at the low-performance end and the `true`[2] BN (2) at the high-performance end.

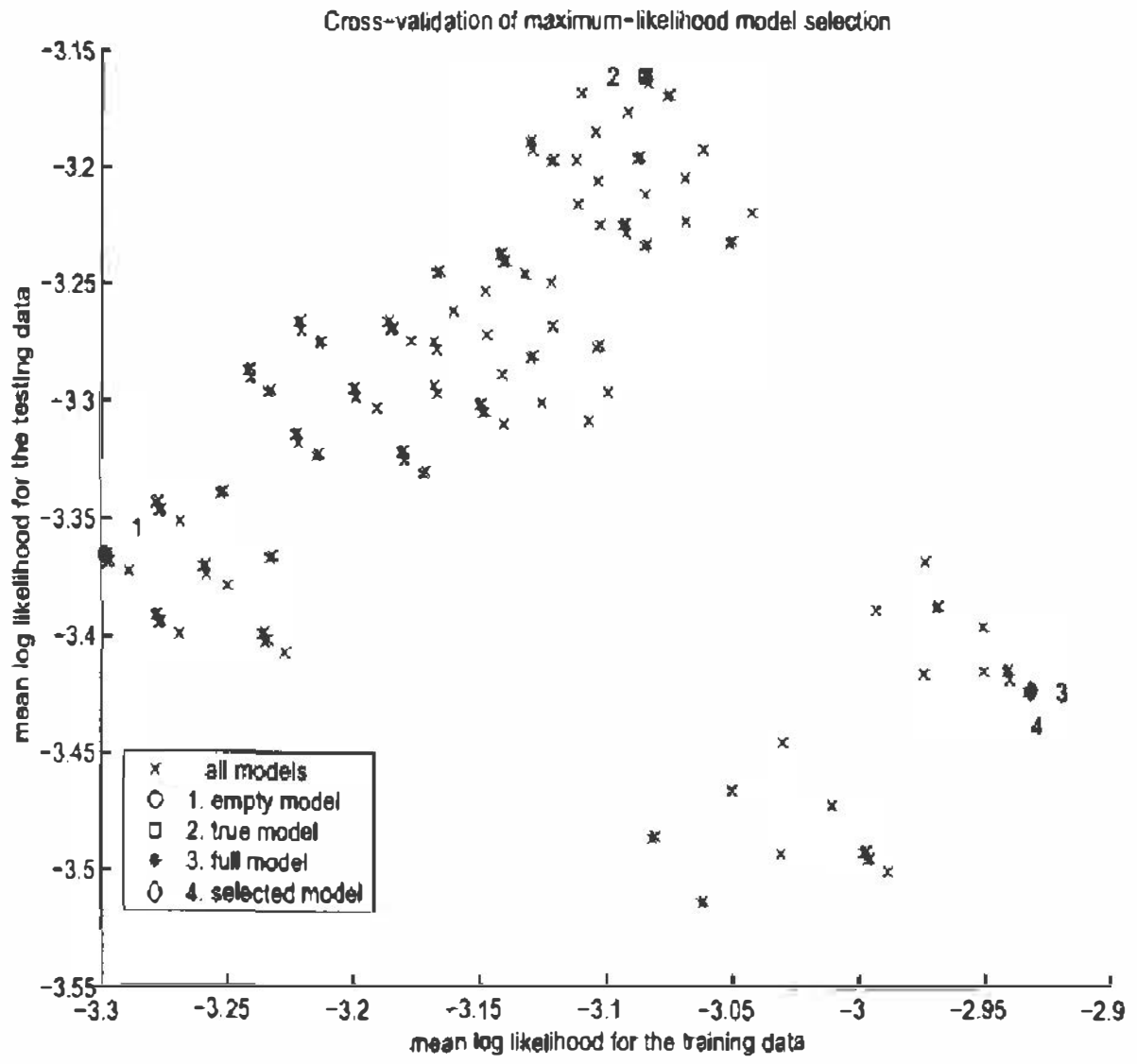

Figure 2: Performances of conventional BN models

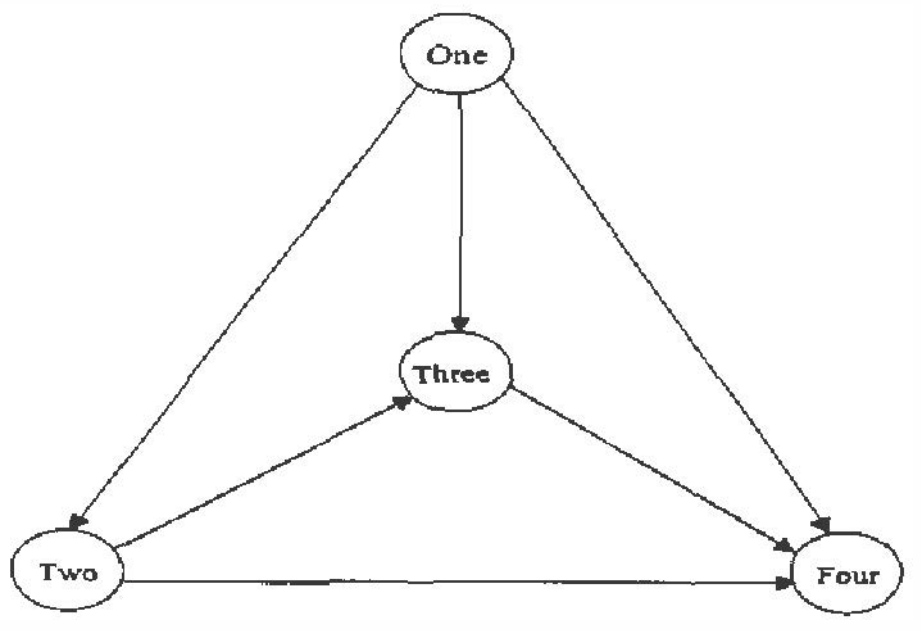

Figure 3: The `full` model structure

For the comparison of the BMN with these conventional BN models, the `full` structure is chosen as the initial reference, with the node ordering `One`≻`Two`≻`Three`≻`Four` chosen for convenience. Since the `full` BN model performs poorly on the testing data, the crucial test for the mixture model concept is to drastically improve its performance. As shown in Figure 4, the performance of the BMN (4) does initially improve towards that of the `true` BN (2), but ultimately degrades to match that of the `full` BN (3) as the number of EM iterations increases. However, by restricting the number of parents that any candidate substructure may have (thus biasing towards the `true` structure), the BMN performance no longer suffers this degradation. In particular, overfitting is prevented by limiting substructures for node `Four` to having only two of the three possible parents. The performance of this restricted BMN (5) then converges closely to that of the `true` BN. Interestingly, the mixture of single-parent substructures (6) also performs quite well, despite its extreme simplicity. However, further restrictions (which bias away from the true structure) degrade the performance towards that of the `empty` BN (1).

[2]There was no significant difference between the `true` data-generating model and the BN with `true` structure and estimated parameters.

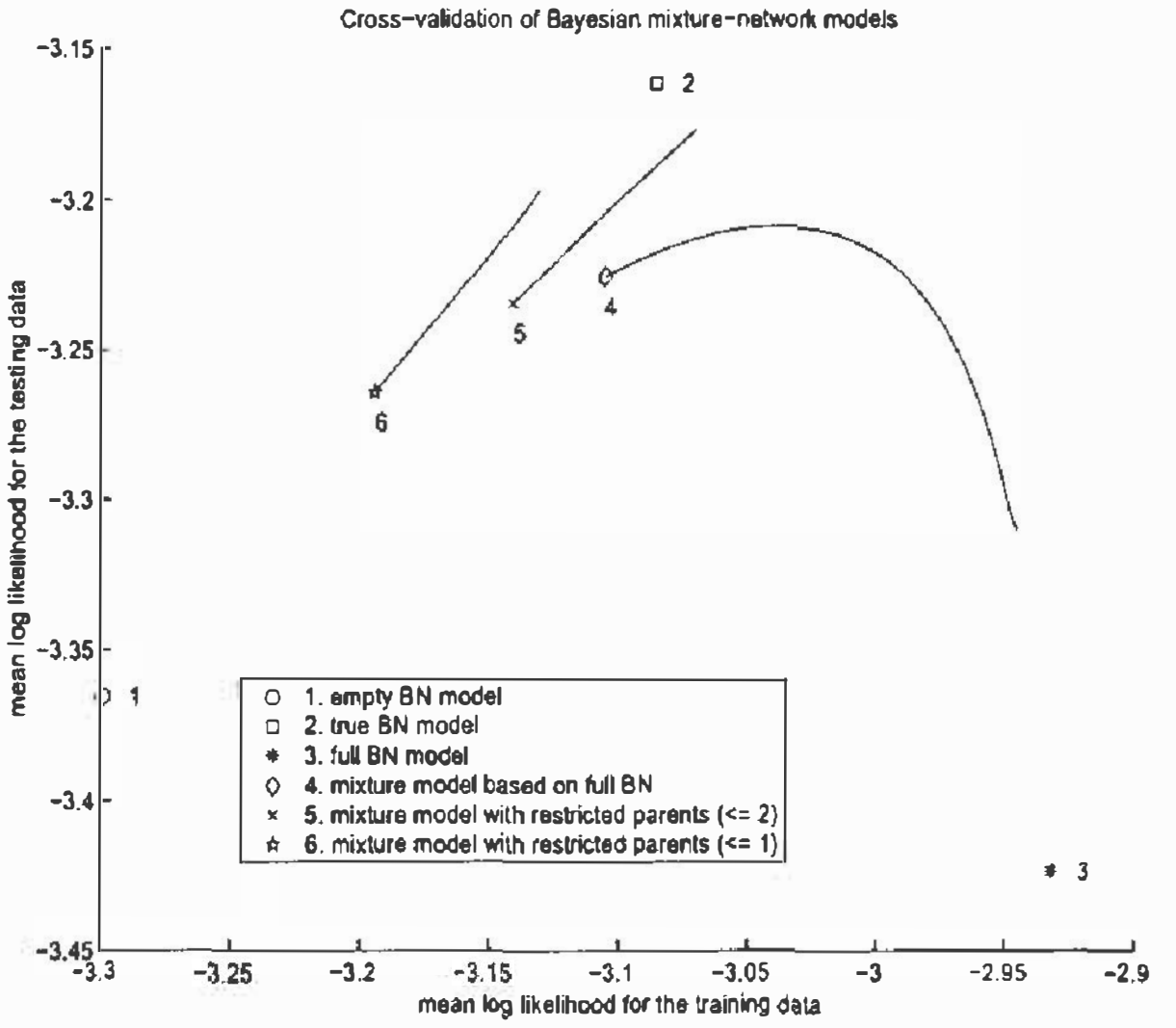

Figure 4: Performance of the BMN model

## 4.2 KEYHOLE PLAN RECOGNITION

In the previous section, the BMN was shown to have improved performance over the conventional BN chosen by model selection for artificial data. Here, the performance of the BMN is analysed on the "real" data observed from an online adventure game known as a *multi-user dungeon* (MUD), in the context of *keyhole*[3] *plan recognition* (Albrecht *et al.* (1997)). The specific task is to predict the current action, location and quest of any player given only knowledge of the player's previous action, location and quest.

The conventional BN models chosen were obtained from the four dynamical Bayesian networks (DBNs) considered by Albrecht *et al.* (1997), shown in Figures 5–8. The main temporal dynamics are encoded in the sequences $A_0, A_1, A_2, \ldots$ of actions and $L_0, L_1, L_2, \ldots$ of locations. The previous quest $Q_0$ and current quest $Q$ remain fixed for each *run* of observations which describe a player completing a single quest. These DBNs were converted to BNs by suitable pre-processing of the data, which included clustering the thousands of possible states to about 40 per node.

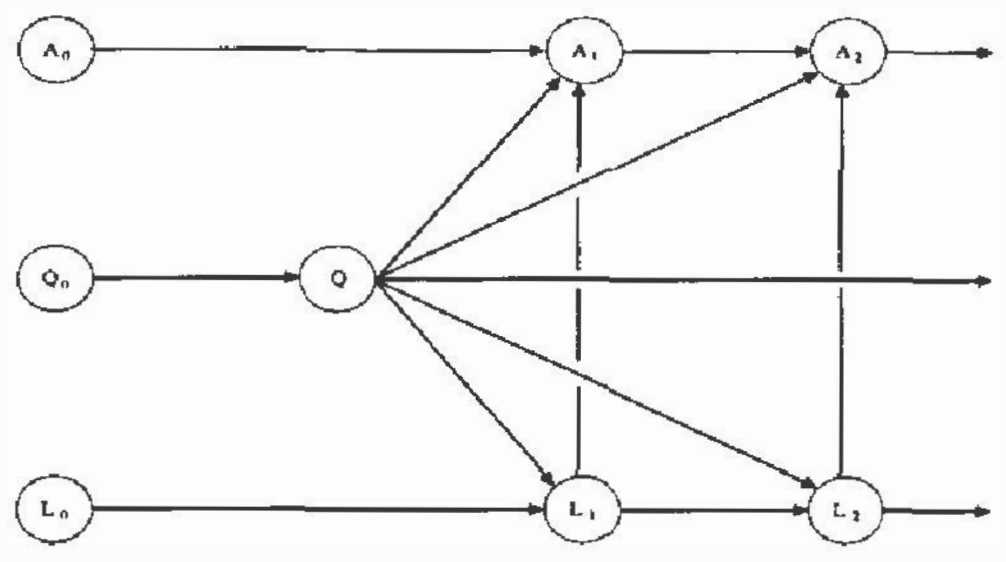

Figure 5: The `main` DBN model

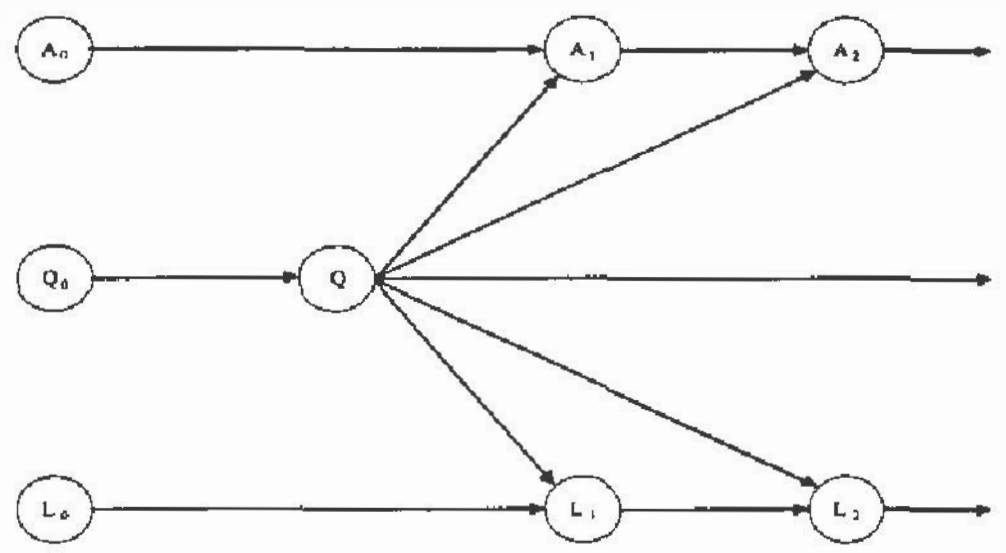

Figure 6: The `independent` DBN model

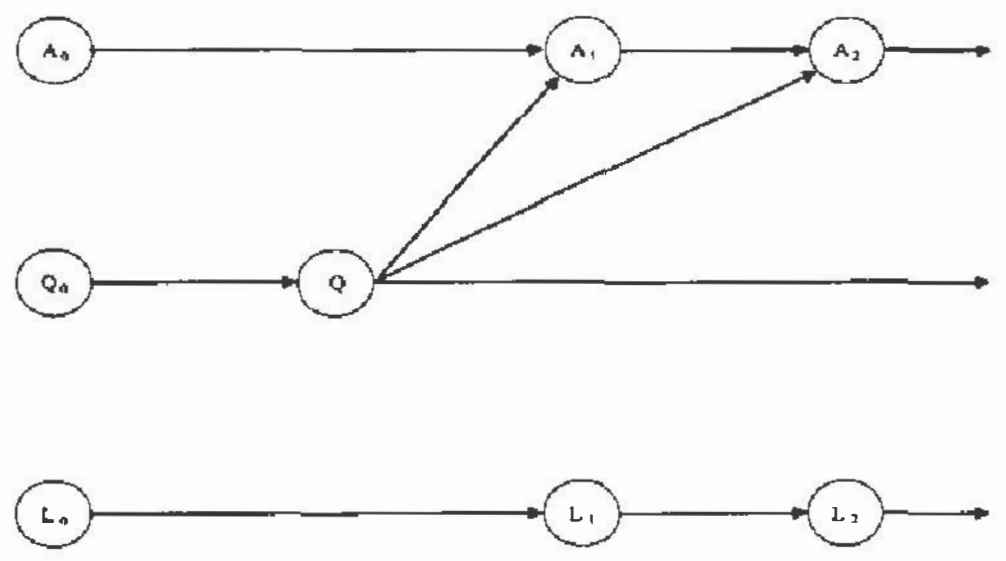

Figure 7: The `action` DBN model

Observe that the `independent` model structure (Figure 6) represents a simplification of the `main` model structure (Figure 5), and that the structures of the `action` (Figure 7) and `location` (Figure 7) models represent further simplifications of the `independent` model structure. Hence, the BMN based on the `main` model structure is an ideal vehicle for comparison with all four models, since it is essentially a mixture of these models (see Section 2.3), plus other models with structures not described here. The mixture of BMN submodels with greatest estimated weights can then be interpreted as the most "likely" global BN model.

The training and testing data used in this experi-ment each comprised 100 randomly sampled *sessions*, in which a player entered the MUD, completed one or more *runs*, and then exited the MUD. The relative performances of both the BMN and the BN models are shown in Figure 9. Observe that the BN models follow a roughly linear trend, with the performances for the testing data improving at the expense of a decreasing fit of the training data. The `main` model (1) overfits the data the most and has the least generalisation. In contrast, the mixture model (6) fits the training data almost as well as the `main` model, but has far greater accuracy for the testing data than all the other models. Interestingly, an examination of the submodel weights for the mixture model reveals that it is effectively an average of the `main` model (with a weight of 0.566) and the `quasi-action` model shown in Figure 10 (weight 0.430). Since the `quasi-action` model (5) by itself has a similar performance to the `main` model (1), the BMN has the property of being much more accurate than the individual models of which it is composed. This is presumably because the resulting CPTs average out the bias of underfitting or overfitting individual submodel CPTs.

[3]The word "keyhole" refers to the fact that observations can only be made from a restricted viewpoint, such that full knowledge is impossible.

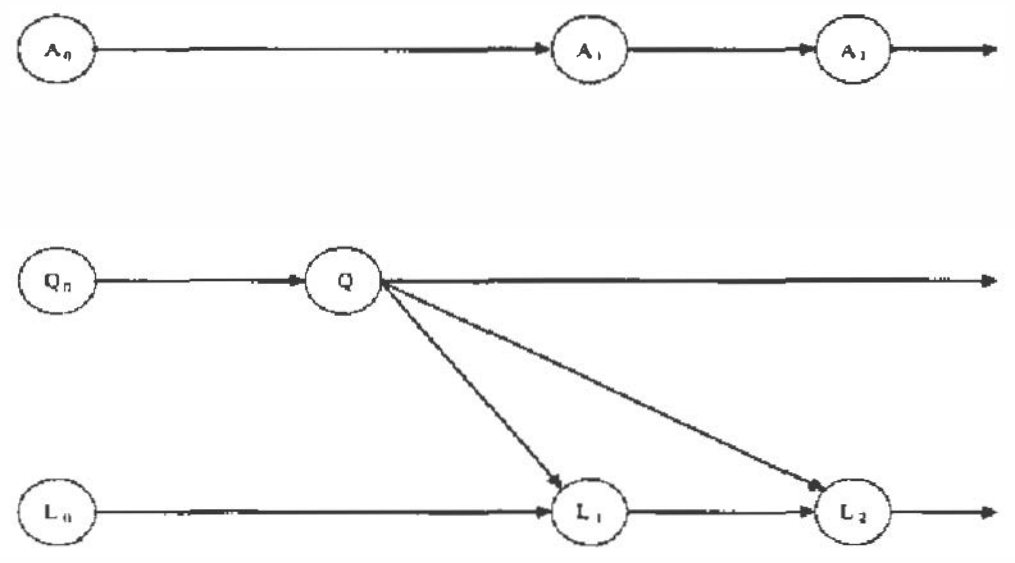

Figure 8: The `location` DBN model

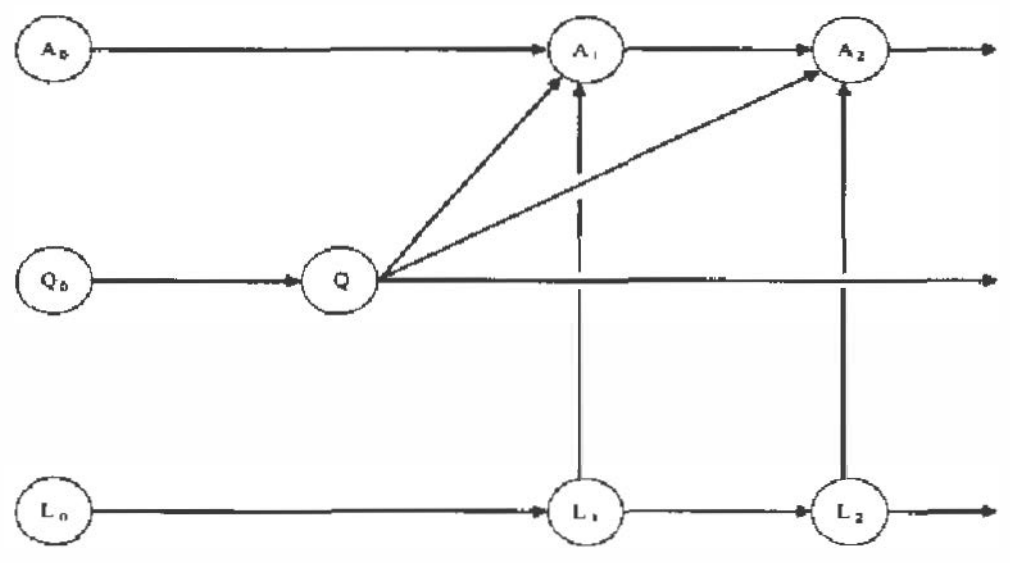

Figure 10: The `quasi-action` DBN model

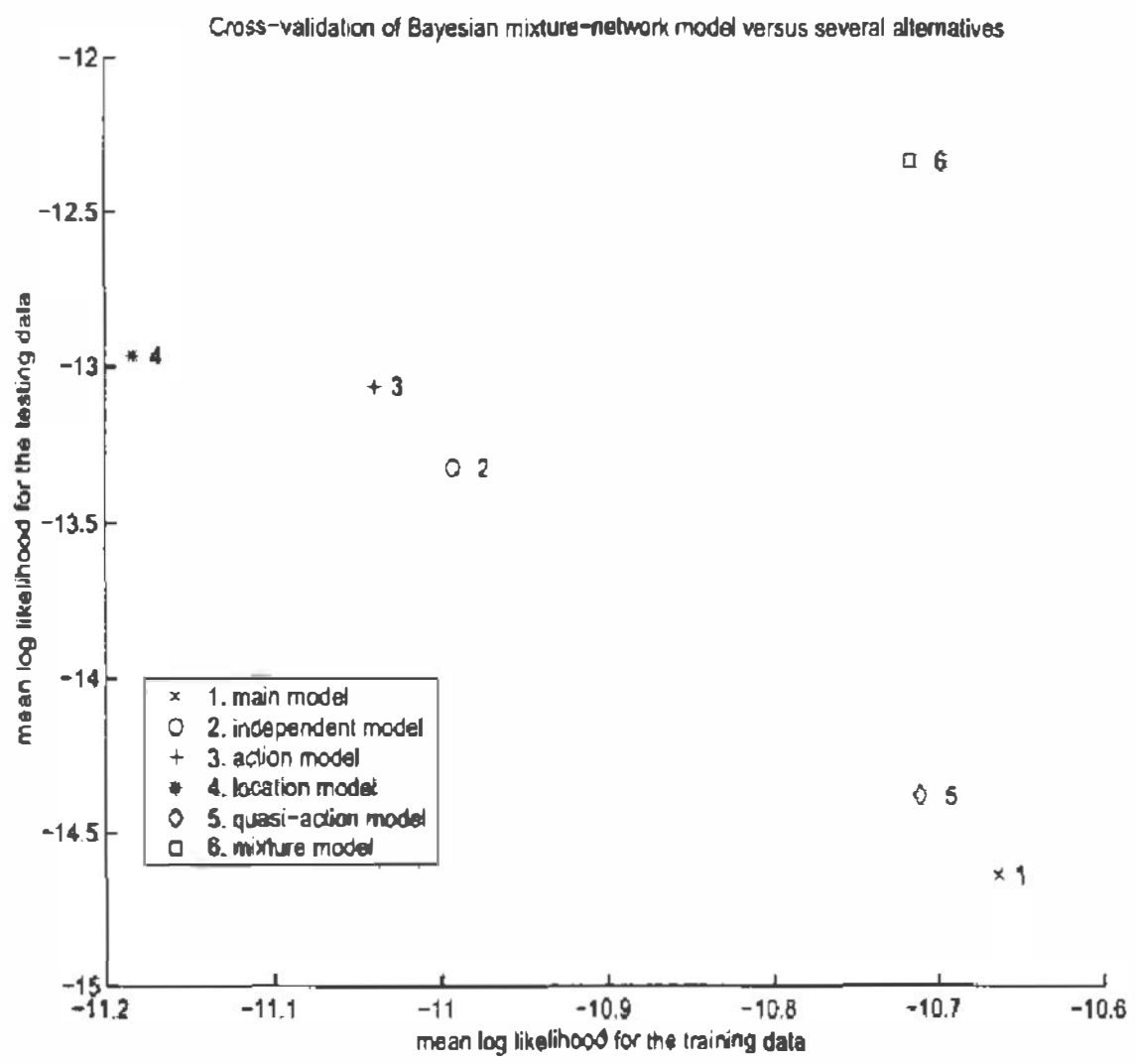

Figure 9: Performance of the BMN on the MUD data compared to conventional BN models

Although the BMN performs very well on simultaneously predicting the current state and the previous state, the results are less clear-cut when predicting individual variables. For instance, the performances of the BMN and conventional BN models in predicting the current quest given the previous quest, action and location are shown in Figure 11. However, for the prediction of the current action given the previous states, the performances shown in Figure 12 reveal that the `location` model (4) is much more accurate than the BMN (6). These mixed results suggest a possible need for conditional BN estimation aimed at optimising the predictions of specific queries (Greiner *et al.* 1997).

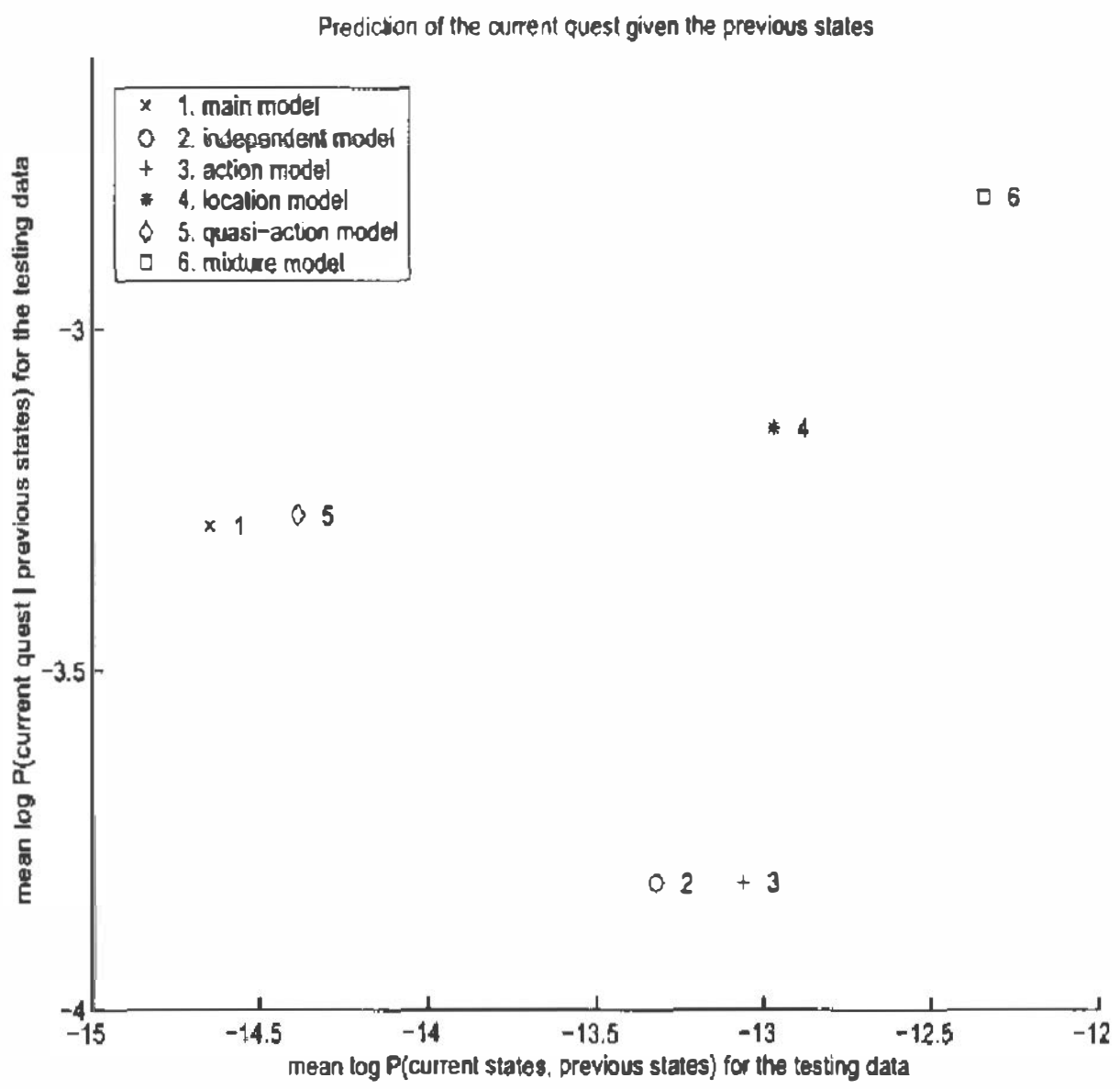

Figure 11: Performances of the BMN and conventional BN models in predicting the current quest

## Acknowledgements

The author wishes to acknowledge the use of the MUD data kindly supplied by David Albrecht of the Department of Computer Science, Monash University, Victoria, Australia.

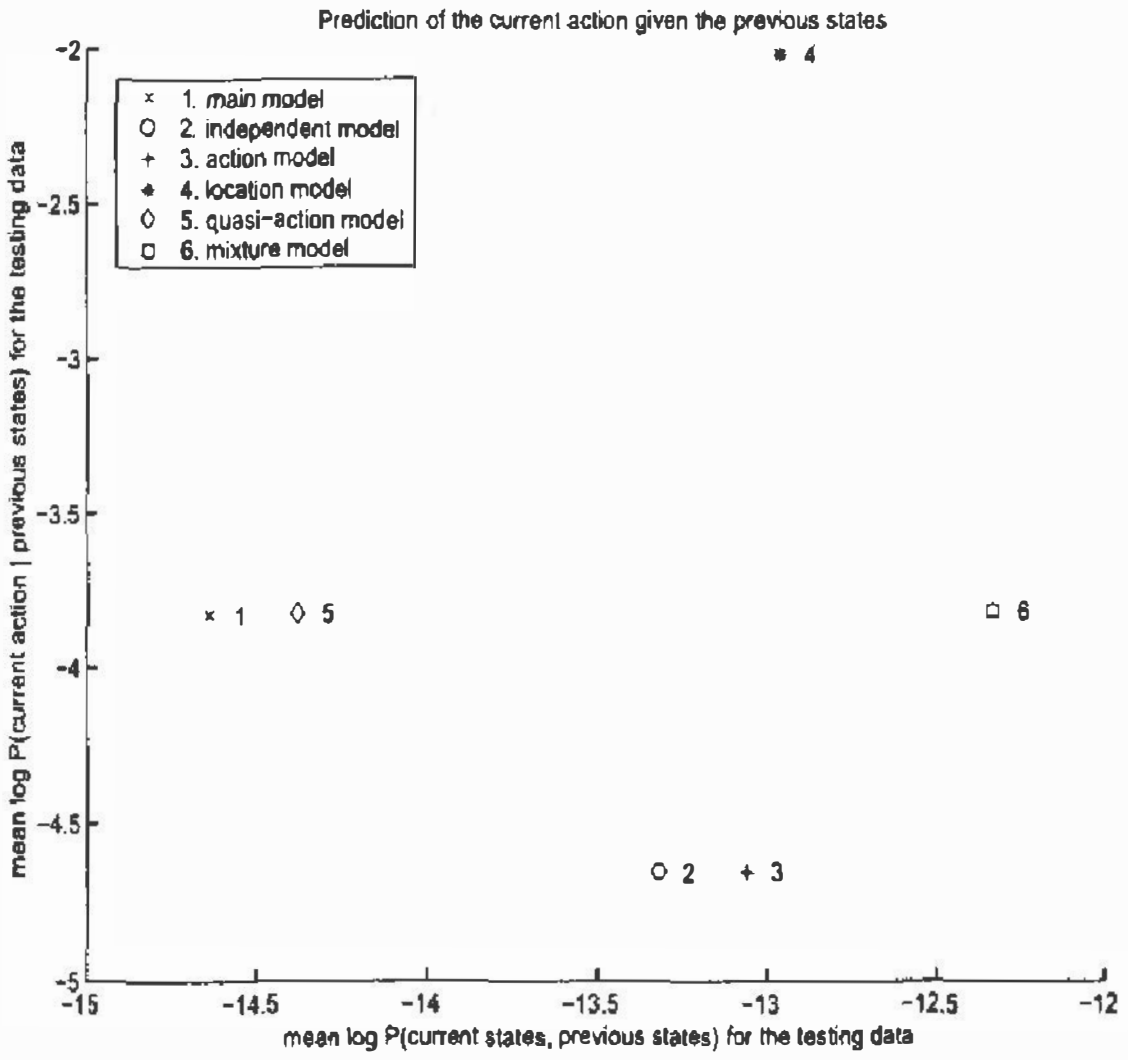

Figure 12: Performances of the BMN and conventional BN models in predicting the current action